\documentclass[sigconf]{acmart}

\usepackage{booktabs} 
\usepackage{balance}

\setcopyright{rightsretained}

\copyrightyear{2018} 
\acmYear{2018} 
\setcopyright{acmcopyright}
\acmConference[KDD '18]{The 24th ACM SIGKDD International Conference on Knowledge Discovery \& Data Mining}{August 19--23, 2018}{London, United Kingdom}
\acmBooktitle{KDD '18: The 24th ACM SIGKDD International Conference on Knowledge Discovery \& Data Mining, August 19--23, 2018, London, United Kingdom}
\acmPrice{15.00}
\acmDOI{10.1145/3219819.3219917}
\acmISBN{978-1-4503-5552-0/18/08}

\begin{document}
\title{Optimization of a SSP's Header Bidding Strategy using Thompson Sampling}

\author[G. Jauvion]{Gr\'{e}goire Jauvion}
\affiliation{%
  \institution{AlephD}
  \city{Paris}
}
\email{gregoire.jauvion@gmail.com}

\author[N. Grislain]{Nicolas Grislain}
\affiliation{%
  \institution{AlephD}
  \city{Paris}
}
\email{ng@alephd.com}

\author[P. Dkengne Sielenou]{Pascal Dkengne Sielenou}
\affiliation{%
  \institution{IMT}
  \city{Toulouse}
}
\email{sielenou_alain@yahoo.fr}

\author[A. Garivier]{Aur\'{e}lien Garivier}
\affiliation{%
  \institution{IMT}
  \city{Toulouse}
}
\email{aurelien.garivier@math.univ-toulouse.fr}

\author[S. Gerchinovitz]{S\'{e}bastien Gerchinovitz}
\affiliation{%
  \institution{IMT}
  \city{Toulouse}
}
\email{sebastien.gerchinovitz@math.univ-toulouse.fr}


\begin{abstract}
Over the last decade, digital media (web or app publishers) generalized the use of real time ad auctions to sell their ad spaces. Multiple auction platforms, also called Supply-Side Platforms (SSP), were created. Because of this multiplicity, publishers started to create competition between SSPs. In this setting, there are two successive auctions: a second price auction in each SSP and a secondary, first price auction, called header bidding auction, between SSPs.

In this paper, we consider an SSP competing with other SSPs for ad spaces. The SSP acts as an intermediary between an advertiser wanting to buy ad spaces and a web publisher wanting to sell its ad spaces, and needs to define a bidding strategy to be able to deliver to the advertisers as many ads as possible while spending as little as possible. The revenue optimization of this SSP can be written as a contextual bandit problem, where the context consists of the information available about the ad opportunity, such as properties of the internet user or of the ad placement.

Using classical multi-armed bandit strategies (such as the original versions of UCB and EXP3) is inefficient in this setting and yields a low convergence speed, as the arms are very correlated. In this paper we design and experiment a version of the Thompson Sampling algorithm that easily takes this correlation into account. We combine this bayesian algorithm with a particle filter, which permits to handle non-stationarity by sequentially estimating the distribution of the highest bid to beat in order to win an auction. We apply this methodology on two real auction datasets, and show that it significantly outperforms more classical approaches.

The strategy defined in this paper is being developed to be deployed on thousands of publishers worldwide.
\end{abstract}

%
%



\maketitle

\section{Introduction}

Real-Time Bidding (RTB) is a mechanism widely used by web publishers to sell their ad inventory through auctions happening in real time. Generally, a publisher sells its inventory through different Supply-Side Platforms (SSPs), which are intermediaries who enable advertisers to bid for ad spaces. A SSP generally runs its own auction between advertisers, and submits the result of the auction to the publisher.

There are several ways for the publisher to interact with multiple SSPs. In the typical ad selling mechanism without header bidding, called the waterfall mechanism, the SSPs sit at different priorities and are configured at different floor prices (typically the higher the priority, the higher the floor price). The ad space is sold to the SSP with the highest priority who bids a price greater than its floor price.

With header bidding, all the SSPs are called simultaneously thanks to a piece of code running in the header of the web page. Then, they compete in a first-price auction which is called the header bidding auction thereafter. In this mechanism, a SSP with a lower priority can purchase the ad if it pays more than a SSP with a higher priority. Consequently, a RTB market with header bidding is more efficient than the waterfall mechanism for the publisher.

\begin{figure}
\centering
\includegraphics[scale=0.3]{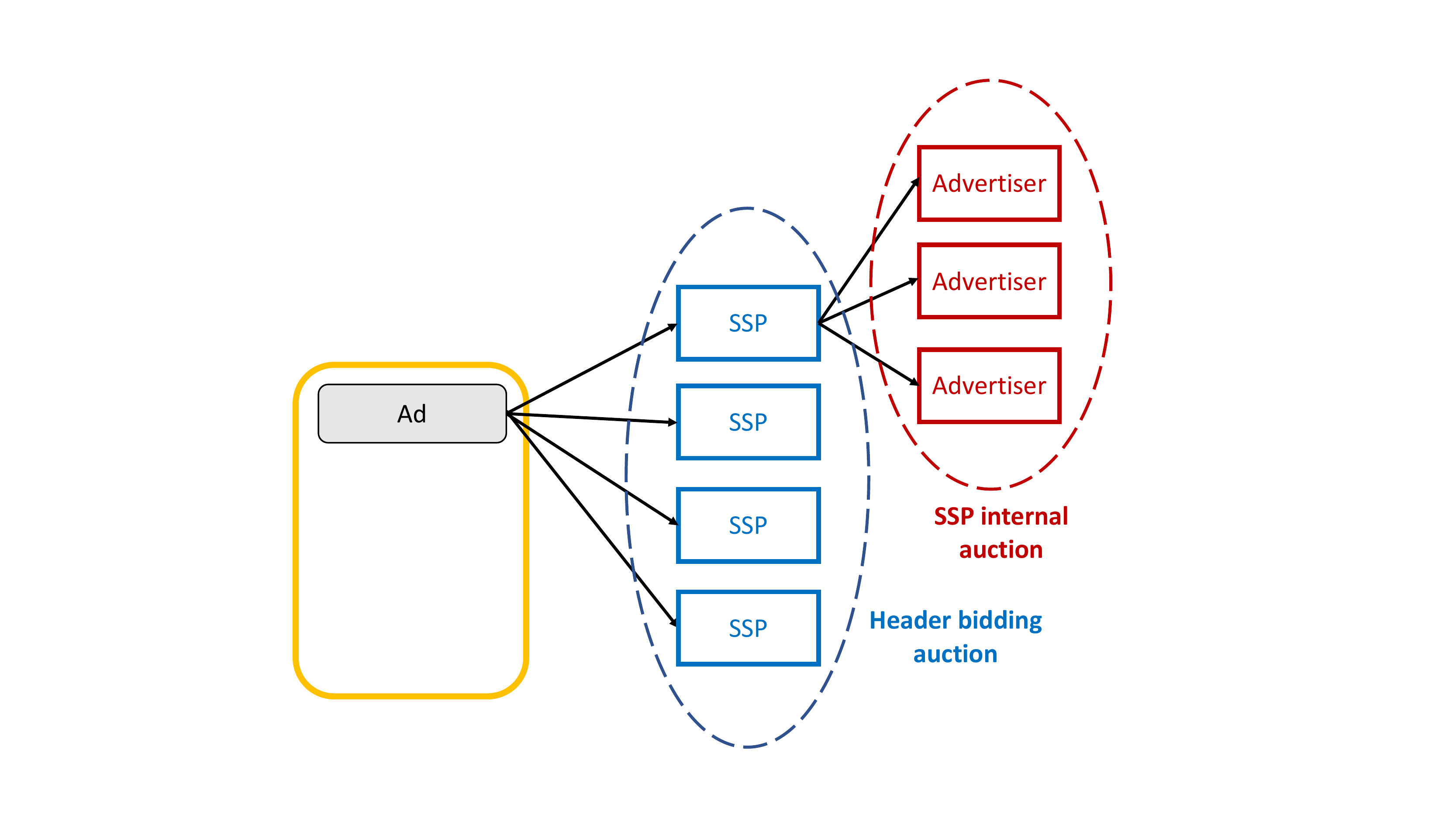}
\caption{Ad-selling process}
Typically some digital content will call a Supply-Side Platform (SSP) when loading, that will itself call many advertisers for bids in a real-time auction.
To increase competition, some publishers have been calling several SSPs to introduce some competition among them.
This setting is called \emph{header bidding} because the competition among SSPs has typically been happening on the client side, in the header of HTML pages.
Header bidding is, in practice, a two-staged process, with several second-price auctions happening in various SSPs, the response of which are aggregated in a final first-price auction.
A SSP may be willing to adjust the bid it is responding to adapt to the first price auction context. This can be seen as an adaptative fee.
\label{ad_selling}
\end{figure}

In this paper, we take the viewpoint of a single SSP buying inventory in a RTB market with header bidding. Based on the result of the auction it runs internally, it submits a bid in the header bidding auction and competes with the other SSPs. This ad-selling process is summarized on Figure \ref{ad_selling}. When it wins the header bidding auction, the SSP is paid by the advertiser displaying its ad, and pays to the publisher the closing price of the header bidding auction. We study the problem of sequentially optimizing the SSP's bids in order to maximize its revenue. Quite importantly we consider a censored setting where the SSP only observes if it has won or lost once the header bidding auction has occurred. The bids of the other SSPs are not observed.

This optimization problem is formalized as a stochastic contextual bandit problem. The context is formed by the information available before the header bidding auction happens, including the result of the SSP internal auction. In each context, the highest bid among the other SSPs in the header bidding auction is modeled with a random variable and is updated in a bayesian fashion using a particle filter. Therefore, the reward (i.e., the revenue of the SSP) is stochastic. We design and experiment a version of the Thompson sampling algorithm in order to optimize bids in a sequence of auctions.

The paper is organized as follows. We discuss earlier works in Section~\ref{relatedworks}. In Section~\ref{pbstatement} we formalize the optimization problem as a stochastic contextual bandit problem. We then describe our version of the Thompson sampling algorithm in Section~\ref{sec:TS}. Finally, in Section~\ref{sec:experiments}, we present our experimental results on two real RTB auction datasets, and show that our method outperforms two more traditional bandit algorithms.

\section{Related work} \label{relatedworks}

\cite{header_bidding} provides a very clear introduction to the header bidding technology, and how it modifies the ad selling process in a RTB market.

Bid optimization has been much studied in the advertising literature. A lot of papers study the problem of optimizing an advertiser bidding strategy in a RTB market, where the advertiser wants to maximize the number of ads it buys as well as a set of performance goals while keeping its spend below a certain budget (see \cite{Wang2017} and references therein). \cite{optimal_bidding} and \cite{optimal_bidding_3} state the problem as a control problem and derive methods to optimize the bid online. In \cite{optimal_bidding_2}, the authors define a functional form for the bid (as a function of the impression characteristics) and write the problem as a constrained optimization problem.

The setting of an intermediary buying a good in an auction and selling it to other buyers (which is what the SSP does in our setting) has been widely studied in the auction theory literature. In \cite{myerson_bilateral}, the author uses the tools developed in \cite{myerson} (one of the most well-known papers in auction theory) to derive an optimal auction design in this setting, and \cite{auctions_intermediaries} and \cite{intermediary_bayesian} analyze how the intermediary should behave to maximize its revenue.

\cite{adx} studies the optimal mechanism a SSP should employ for selling ads and analyzes optimal selling strategies. \cite{adx_header_bidding} analyzes the optimal behaviour of a SSP in a market with header bidding, and validates the approach on randomly generated auctions.

From an algorithmic perspective, the Thompson sampling algorithm was introduced in \cite{thompson}. The papers \cite{Kaufmann2012,Kaufmann2013} studied its theoretical guarantees in parametric environments, while \cite{Leike2016} studied it in non-parametric environments. Besides, a very clear overview of the particle filtering approach to update the posterior distribution is given in \cite{Doucet2001,murphy}.

Bandit algorithms were already designed and studied for repeated auctions, including RTB auctions. For instance, in repeated second-price auctions, \cite{WePeRi-16colt-RepeatedAuctions} construct a bandit algorithm to optimize a given bidder's revenue, while \cite{CeGeMa-15-ReservePriceOptimization} design a bandit algorithm to optimize the seller's reserve price. 

In a setting very similar to ours, \cite{HeidariETAL-16-PricingLowRegretSeller} study the situation where a given SSP competes with other SSPs in order to buy an ad space. They design an algorithm that provably enables the SSP to win most of the auctions while only paying a little more than the expected highest price of the other SSPs. Though the problem seems similar, our objective is different: we want the SSP to maximize its revenue, and not necessarily to win most auctions with a small extra-payment. In particular we cannot neglect the closing price of the SSP's internal auction in the optimization process.

We finally mention the work of \cite{KlLe-03-PostedPriceAuctions} for the online posted-price auction: for each good in a sequence of identical goods, a seller chooses and announces a price to a new buyer, who buys the good provided the price does not exceed their private valuation (see also \cite{MoMe-14nips-PostedPriceAuctions,MoMe-15nips-PostedPriceAuctions-randomvaluation} when the seller faces strategic buyers). Though their problem is different, the shape of their reward function is very similar to ours. The authors show that the classical UCB1 and Exp3 bandit algorithms applied to discretized prices are worst-case optimal under several assumptions on the sequence of the buyers' valuations. In our paper we do not tackle the worst case and instead use prior knowledge on the ad auction datasets (i.e., an empirically-validated parametric model) to better optimize the SSP's revenue.

\section{Problem statement}\label{pbstatement}

\subsection{RTB market}

We represent the RTB market as an infinite sequence $\mathcal{D}$ of time-ordered impressions $1,\ldots,n,\ldots$ happening at times $t_1,\ldots,t_n,\ldots$. We note $\mathcal{D}_t$ the sequence of impressions happening before time $t$ (including $t$).

Impression $i$ happening at time $t_i$ is characterized by a context $c_i$, which summarizes all the information
relative to impression $i$ that is available before the header bidding auction starts. It may contain the ad placement (where it is located on the web page), some properties of the internet user (for example its operating system), or the time of the day. An important variable of the context which is specific to our setting is the closing price of the SSP internal auction, which is known before the header bidding auction happens.

We assume that the context is categorical with a finite number of categories $C$. A continuous variable can be discretized to meet this assumption. Without loss of generality, we assume that the categories are $1,\ldots,c,\ldots,C$. We note $\mathcal{D}_{t,c}$ the subsequence of $\mathcal{D}_t$ containing all impressions $i$ such that $c_i=c$.

\subsection{Ad selling process with header bidding}

We assume that $S$ SSPs: $\mathcal{S}_1,\ldots,\mathcal{S}_S$ compete in the header bidding auctions (possibly bidding $0$ if they are not interested in purchasing the ad). We note $b_{i,s}$ the bid of $\mathcal{S}_s$ in the header bidding auction for impression $i$. As the header bidding auction is a first-price auction, its closing price is $\max_s(b_{i,s})$.

From now on, we consider the problem from $\mathcal{S}_1$ standpoint. We note $q_i = b_{i,1}$ the bid submitted by $\mathcal{S}_1$ in the header bidding auction, which is the variable to optimize. We also write $x_i = \max(b_{i,2},\ldots,b_{i,S})$, which is the highest bid among the other SSPs.

In each impression $i$, we assume that $\mathcal{S}_1$ runs an internal auction between advertisers, whose closing price is denoted $p_i$. $p_i$ is the amount paid by the advertiser winning the internal auction to $\mathcal{S}_1$ should $\mathcal{S}_1$ win the header bidding auction. Note that we do not need to know the detailed internal auction mechanism but only its closing price.

Before header-bidding, a SSP would run a second-price auction with an advertiser bidding \$10 and closing at $p_i = \$8$. Then the SSP would respond $q_i = p_i - \text{fees} = \$6$ to the publisher. In this context, the advertiser pays $\$8$, the publisher receives $\$6$ and the SSP gets its fees: $\$2$. In a header bidding context, the SSP is in competition with other SSPs in a first price auction, it may lose an opportunity by taking too much fees or pay too much if it is sure to win and take too little fees. 

\subsection{Revenue function for the SSP $\mathcal{S}_1$}

The revenue function $R_i(.)$ of $\mathcal{S}_1$ at impression $i$ can be written as $R_i(q) = \mathbf{1}_{q \geq x_i}(p_i - q)$. Indeed:
\begin{itemize}
\item When $q \geq x_i$, $\mathcal{S}_1$ wins the header bidding auction. It is paid $p_i$ by the advertiser winning the internal auction, and it pays $q$ to the publisher.
\item Otherwise, $\mathcal{S}_1$ does not display any ad and gets no revenue in the auction.
\end{itemize}

\begin{figure}
\centering
\includegraphics[scale=0.4]{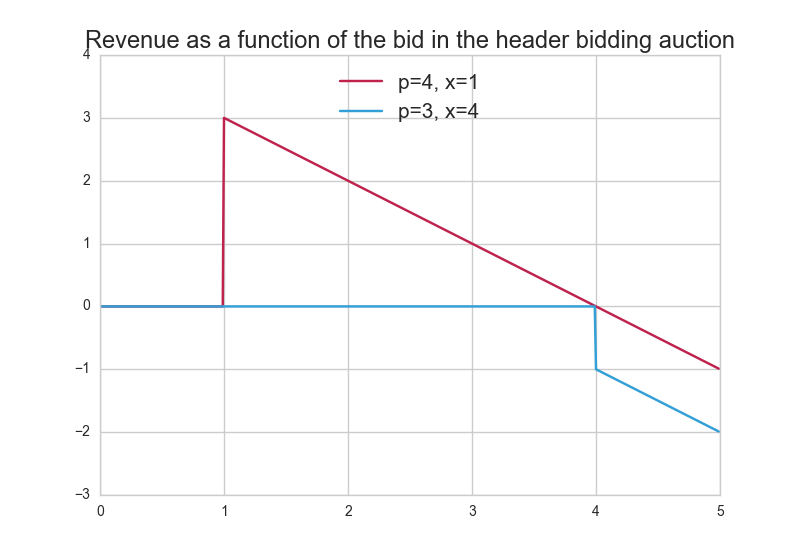}
\caption{$\mathcal{S}_1$ revenue as a function of its bid $q$}
\label{revenue}
\end{figure}

In Figure \ref{revenue} we plot $\mathcal{S}_1$'s revenue as a function of its bid $q$, for two sets of values for the closing price $p$ of the internal auction and the highest bid $x$ among the other SSPs.

Note that in the setting described here, we ignore some factors having an impact on $\mathcal{S}_1$'s revenue. Indeed $\mathcal{S}_1$ may charge a fee to the advertiser in addition to the closing price of the internal auction it runs. Also, the cost of running the internal auction may lower $\mathcal{S}_1$'s revenue. These factors would impact the revenue function but the strategy described in this paper would remain applicable.

\subsection{Optimization problem statement}

Before the header bidding auction for impression $i$ happens, the value $x_i$ of the highest bid among the other SSPs is unknown and is modeled with the random variable $X_i$. Thus, the revenue optimization problem over $n$ impressions can be expressed as follows:
\begin{equation}
\begin{split}
& \max_{q_1,\ldots,q_n} \mathbf{E}\left[\sum_{i=1}^n \mathbf{1}_{q_i \geq X_i}(p_i - q_i)] \right] \\
& \qquad = \max_{q_1,\ldots,q_n} \sum_{c=1}^C \sum_{i \in \mathcal{D}_{t_n,c}} \mathbf{E}[\mathbf{1}_{q_i \geq X_i}(p_i - q_i)] \,,
\end{split}
\end{equation}
where the maximum is over the prices $q_1,\ldots,q_n$ that the SSP can choose as a function of the past observations.

It would be tempting to model the variables $X_i$ as independent and identically distributed within any context $c$, with an unknown distribution $\phi_c$. Under this assumption, the task presented here boils down to a contextual stochastic bandit problem.
A closer look at the data, however, shows that there are significant non-stationarities in time. We explain below that our final model does address this issue, by the use of a particle filter within the bandit algorithm. 

We emphasize that, after the header bidding auction for impression $i$ has occurred, $\mathcal{S}_1$ does not observe the value of the bid $x_i$, but only observes if it has won or lost the header bidding auction, i.e., $\mathbf{1}_{q_i \geq x_i}$. This censorship issue must be tackled in the optimization methodology.

\section{Revenue optimization using Thompson sampling} \label{sec:TS}

In this section we present a method to sequentially optimize the bids~$q_i$. It combines the Thompson sampling algorithm with a parametric model for the distributions $\phi_c$ (recall that $\phi_c$ is the distribution of the other SSPs' highest bids $X_i$ within context $c$). Note that all the contexts $c$ are modeled independently.

\subsection{Parametric estimation of $\phi_c$}

We introduce $f_{\theta}$ a family of distributions parametrized with $\theta$, and we note $F_{\theta}$ the corresponding cumulative density functions. For each context $c$, we assume that the distribution $\phi_c$ of the other SSPs' highest bids $X_i$ belongs to the family $f_{\theta}$; let $\theta_c$ be such that $\phi_c = f_{\theta_c}$.

According to the Thompson sampling method, we fix a prior distribution $\pi_{c,0}(\theta)$ over $\theta_c$. Then, for all $t$, we consider the posterior distribution $\pi_{c,t}(\theta)$ given all the observations available at the end of the $t$-th auction, i.e., the censored observations $\mathbf{1}_{q_i \geq x_i}$ for $i  \in \mathcal{D}_{t,c}$.

The Bayes rule yields the following expression for $\pi_{c,t}$:

\begin{equation}
\pi_{c,t}(\theta) \propto \pi_{c,0}(\theta) \prod_{i \in \mathcal{D}_{t,c}} \big[F_{\theta}(q_i) \mathbf{1}_{x_i \leq q_i} + (1 - F_{\theta}(q_i)) \mathbf{1}_{x_i > q_i}\big]
\end{equation}

\subsection{Overview of the methodology}

In our model the Thompson sampling algorithm unfolds as follows: before any impression $i$,
\begin{itemize}
\item Sample a value $\theta$ from the posterior distribution $\pi_{c_i,t_{i-1}}$;
\item Compute the bid $q_i$ that would maximize the SSP's expected revenue if $X_i \sim f_{\theta}$ (see below);
\item Observe the auction outcome $\mathbf{1}_{q_i \geq x_i}$ and update the posterior $\pi_{c_i,t_{i}}$.
\end{itemize}

As the particle filter provides a discrete approximation of the posterior distribution, the sampling step is straightforward. 
The optimization of the bid $q_i$ is a one-dimensional optimization problem: when $X_i \sim f_{\theta}$, the maximal SSP expected revenue is 
$$\max_q (p_i - q) F_{\theta}(q)\;.$$
There is no closed form solution in general, but this problem can be solved numerically for example by using Newton's method.

The difficult step of the algorithm is the update of the posterior, which is explained in the next section. 

\subsection{Updating the posterior distribution}\label{posterior}

It would be difficult to sample directly from the posterior distribution $\pi_{c,t}$, which does not have a simple or tractable form. Even the use of MCMC methods like Metropolis-Hastings would be hazardous, since computing the density of the posterior distribution has a linear cost in the number of past observations which is huge in advertising \footnote{Indeed, the profile of the payoff function induces a posterior distribution that cannot be simplified. Hence, computing the posterior density exactly, cannot be done better than by computing the product of all bayesian updates, which in practice is intractable and rules-out MCMC sampling.}.

To overcome these difficulties, we approximate the posterior distribution with a \emph{particle filter}, a powerful sequential Monte-Carlo method for Hidden Markov Models (HMM). For an introduction on HMM and particle filtering, we refer to \cite{Cappe05}. The basic idea of a particle filter is to approximate the sequence of posterior distributions by a sequence of discrete probability distributions which are derived from one another by an \emph{evolution} procedure (which may include a \emph{selection} step). The posterior distribution $\pi_{c,t}$ is estimated by a discrete distribution on $K$ points called \emph{particles}. The particles are denoted by $(\theta_{c,1,t},\ldots,\theta_{c,K,t})$ and their respective weights by $(w_{c,1,t},\ldots,w_{c,K,t})$. The evolution procedure and the selection step we use are described below. 

A very important strength of the particle filter approach is that it allows to handle \emph{non-stationarity}: the HMM model encompasses the possibility that the hidden variable (here, the unknown parameter $\theta$) evolves in time according to a Markovian dynamic of kernel $p(\theta'|\theta)$, thus forming an unobserved sequence $(\theta_t)_t$. We use this possibility by assuming that the parameter $\theta_t$ is equal to $\theta_{t-1}$ plus a small step in an unknown direction: this permits to handle parameter drift directly inside of the model. 

The theory of particle filters for general state space HMM \cite{Crisan2000,Doucet2001,Crisan2002,Douc2012} suggests that, in cases such as ours, the particle approximations converge to the true posterior distributions of the parameter $\theta$ when the number of particles tends to infinity.

\subsubsection{Evolution: updating the distribution}

Recall that we run $C$ independent instances of Thompson Sampling, one for each context~$c$. Next we focus on one context $c$ and recall how to update the particle distribution in the particle filter. To simplify the notation, we write $t-1$ and $t$ for the times of two consecutive impressions within context $c$, even if other contexts appeared in between.

The update consists of two steps. First the particles $\theta_{c,k,t}$ are sampled from a proposal distribution $q(\theta_{c,k,t}|\theta_{c,k,t-1},\mathbf{1}_{q_t \geq x_t})$. We then compute new unnormalized weights $\hat{w}_{c,k,t}$ by importance sampling:
\begin{equation}
\begin{split}
\hat{w}_{c,k,t} = \hat{w}_{c,k,t-1} & \times \big[F_{\theta_{c,k,t}}(q_t) \mathbf{1}_{x_t \leq q_t} + (1 - F_{\theta_{c,k,t}}(q_t)) \mathbf{1}_{x_t > q_t}\big] \\
& \times \frac{p(\theta_{c,k,t}|\theta_{c,k,t-1})}{q(\theta_{c,k,t}|\theta_{c,k,t-1},\mathbf{1}_{q_t \geq x_t})} \;,
\end{split}
\end{equation}
where $p(\theta'|\theta)$ is the transition kernel of the hidden process. 
Here, we may simply take the proposal distribution $q(\theta_{c,k,t}|\theta_{c,k,t-1},\mathbf{1}_{q_t \geq x_t})$ to be equal to the transition distribution $p(\theta_{c,k,t}|\theta_{c,k,t-1})$, which yields:
\begin{equation}
\hat{w}_{c,k,t} = \hat{w}_{c,k,t-1} \times \big[F_{\theta_{c,k,t}}(q_t) \mathbf{1}_{x_t \leq q_t} + (1 - F_{\theta_{c,k,t}}(q_t)) \mathbf{1}_{x_t > q_t}\big] \,.
\end{equation}

\noindent
The normalized weights $w_{c,k,t}$ can be computed as:
$$w_{c,k,t} = \frac{\hat{w}_{c,k,t}}{\sum_{k'=1}^K \hat{w}_{c,k',t}} \;.$$

\subsubsection{Selection: resampling step}

The basic update described previously generally fails after a few steps because of a well-known and general problem: weight degeneracy. Indeed, most of the particles soon get a negligible probability, and the discrete approximation becomes very poor.
A standard strategy used to tackle this issue is the use of a resampling step when the degree of degeneracy is considered to be too high.
We use the following methodology given in \cite{murphy} for resampling:
\begin{itemize}
\item Compute $S = \Big(\sum_{k=1}^K w_{c,k,t}^2\big)^{-1}$ to quantify the degree of degeneracy of the particle filter
\item If $S < S_{\min}$ ($S_{\min}$ is a hyperparameter of the particle filter), resample all the particles by sampling $K$ times with replacement the current set of weighted particles $\left\{\theta_{c,1,t},\ldots,\theta_{c,K,t}\right\}$. The result is an unweighted sample of $K$ particles, so we set the new  weights to $\hat{w}_{c,k,t} = \frac{1}{K}.$
\end{itemize}

There exist some alternative resampling schemes that could be used: see~\cite{Douc2012} for a presentation of some of them, and for a discussion on their convergence properties and computation cost. 

\ \\
\textit{Time and space complexities.} Recall that $K$ is the number of particles and that $C$ is the number of contexts. After each new impression~$i$, since $i$ only falls into one context $c$, the evolution and selection steps described above need only be carried out for this particular $c$. This thus only requires $\mathcal{O}(K)$ elementary operations per impression (including calls to the cumulative distribution function $F_{\theta}(x)$). As for space complexity, a direct upper bound is $\mathcal{O}(C K)$ since we need to store weight vectors for each context $c=1,\ldots,C$.

\subsection{Implementation of Thompson sampling} \label{models}
We may now detail our modelling and algorithmic choices for the particle filter within the Thompson sampling algorithm.

\paragraph{Distribution of the highest bids $x_i$.} We model the highest bids $x_i$ among the other SSPs with a lognormal distribution, a standard choice in econometrics or finance. Lognormal distributions are parametrized by $\theta=\left(\theta^{(1)}, \theta^{(2)}  \right),$ where $\theta^{(1)}=\sigma>0$ and $\theta^{(2)}=\mu \in \mathbb{R}.$ Here, $\mu$ and $\sigma$ are respectively location and scale parameters for the normally distributed logarithm $\ln(x_i).$

\paragraph{Particle filter.}
We write $\theta_c$ for the parameters of the lognormal distribution associated with context $c$. The particle filter for the posterior distributions works as follows:
\begin{enumerate}

\item In order to handle non-stationarity, we model the parameters $\theta_{c}$ by Markov chains  $\theta_{c,t}$ such that 
$\log\left(\theta^{(1)}_{c,t}\right)=\log\left(\theta^{(1)}_{c,t-1}\right)+ E_1$ and
$\theta^{(2)}_{c,t}=\theta^{(2)}_{c,t-1}+ E_2$, where $E_1,E_2\sim \mathcal{N}(0,\epsilon=0.005)$ are independent Gaussian variables with mean $0$ and standard deviation $\epsilon.$

\item At each time $t$, for each context $c\in\{1,\dots,C=100\}$, we use $K=100$ particles $\theta_{c,k,t}=\left(\theta^{(1)}_{c,k,t}, \theta^{(2)}_{c,k,t}  \right).$

\item As explained above, the particles $\theta_{c,k}$ evolve at step $t$ according to the same dynamic as the unobserved parameters $\theta_{c,t}$: 
$\log\left(\theta^{(1)}_{c,k,t}\right)=\log\left(\theta^{(1)}_{c,k,t-1}\right)+\mathcal{N}(0,\epsilon=0.005)$ and
$\theta^{(2)}_{c,k,t}=\theta^{(2)}_{c,k,t-1}+\mathcal{N}(0,\epsilon=0.005).$

\item We use a uniform distribution as prior $\pi_{c,k,0}(\theta)$ for the parameter $\theta_{c,0}$, and thus uniformly generate the components of the initial particle $\theta_{c,k,0}=\left(\theta^{(1)}_{c,k,0}, \theta^{(2)}_{c,k,0}  \right).$ Because of the high number of auctions in each context, the choice of the prior distribution $\pi_{c,k,0}(\theta)$ has little impact on the result, as long as its support contains the parameter $\theta_c$.

\item Finally, we choose $S_{min}=K/2$ as a resampling threshold criterion.

\end{enumerate}

\section{Experiments on RTB auctions datasets} \label{sec:experiments}

\subsection{Constrution of the datasets}\label{data}

In practice, the SSPs generally do not share their bids with one another, and we do not have a dataset with the bids from all SSPs in header bidding auctions. The datasets we have used in these experiments give, for two web publishers, the bids as well as the names of the advertisers in RTB auctions run by a particular SSP over one week, in a setting without header bidding.

For these two web publishers, a dataset giving both the bids in $\mathcal{S}_1$ internal auction and the bids from other SSPs in the header bidding auction has been artificially built the following way:
\begin{itemize}
	\item All the advertisers competing in the RTB auctions (typically a few dozens) have been randomly assigned to one of two groups of advertisers named A and B
	\item In each auction, the bids coming from advertisers in the group A are supposed to be the bids of the internal auction run by the SSP $\mathcal{S}_1$, and the bids coming from advertisers in the group B are supposed to be the bids coming from the other SSPs in the header bidding auction
	\item Hence, in a given auction $i$, the closing price of the internal auction $p_i$ is given by the second highest bid from advertisers in the group A, and the highest bid among other SSPs $x_i$ is given by the highest bid from advertisers in the group B
	\item The auctions where there are less than two bids from advertisers in the group A or less than one bid from advertisers in the group B have been removed from the dataset
\end{itemize}

These two datasets are named $P_1$ and $P_2$ thereafter. A brief description is given in Table~\ref{tab:publishers}. We give the share of auctions where $x_i\leq p_i$, which is the share of auctions where the SSP $\mathcal{S}_1$ could have won the header bidding auction while generating a positive revenue, by choosing $q_i \in [x_i,p_i]$.

\begin{table}
\caption{Some properties of the datasets $P_1$ and $P_2$.}
\begin{minipage}{\columnwidth}
\begin{center}
\label{tab:publishers}
\begin{tabular}{@{}lcc@{}}
\toprule
& $P_1$ & $P_2$ \\
\midrule Number of auctions & 1,496,294 & 410,840 \\
 Number of users  & 875,634 & 269,272 \\
    Number of ad placements & 3,526 & 31 \\
    Share of auctions where $x_i\leq p_i$ & $55.2\% $ & $48.4\% $ \\
\bottomrule
\end{tabular}
\end{center}
\end{minipage}
\end{table}

The experiments have been performed in the two following configurations:
\begin{itemize}
	\item Stationary environment: the data is shuffled. This configuration is used to evaluate the strategy in a stationary environment
	\item Non-stationary environment: the data is sorted in chronological order. In this case, the data is non-stationary, as the bids highly depend on the time of the day. This configuration is used to evaluate the strategy in a non-stationary environment
\end{itemize}

Note that all the bids have been multiplied by a constant.

\subsection{Definition of the contexts}\label{contexts}

In the experiments, we define the context in auction $i$ by the closing price of $\mathcal{S}_1$ internal auction $p_i$. The closing price $p_i$ is transformed into a categorical context by discretizing it into $C$ disjoint bins.

The $l$-th bin contains all auctions where $p_i \in \left[ q(\frac{l-1}{C}), q(\frac{l}{C}) \right[$, where $q$ is the empirical quantile function of the closing prices $(p_i)$ estimated on the data. Consequently, each one of the $C$ contexts contains approximately the same number of auctions.

The number of contexts should be chosen carefully. A high number of contexts enables to model more precisely the distribution of the highest bid among other SSPs, which is modeled independently on each context, at the price of a slower convergence. In the experiments, we have chosen $C=100$ which yields a good performance on the datasets.

\subsection{Baseline strategies}\label{models2}

We define in this section the baseline strategies used to assess the quality of the Thompson sampling strategy. They correspond to the use of classical multi-armed bandit (MAB) models~\cite{PLG06}. Each arm $j=1,\ldots,J$ corresponds to a coefficient $\alpha^{(j)}=\frac{j}{J}$ applied to the closing price of the internal auction $p_i$ to obtain the bid of the SSP $\mathcal{S}_1$, $q_i = \alpha^{(j)} \cdot p_i$. Note that this strategy implies that $q_i \leq p_i$, as the revenue for the SSP $\mathcal{S}_1$ can not be positive when $q_i > p_i$.

In each auction $i$, the SSP $\mathcal{S}_1$ chooses an arm $j(i)$ and bids $q_i=\alpha^{j(i)} \cdot p_i$. Then, it receives a reward equal to $\mathbf{1}_{q_i \geq x_i}(p_i - q_i)$, and the rewards associated to the other arms are unknown.

The goal of the SSP is to maximise their expected cumulative reward. In the MAB literature, this reward maximisation is typically defined via the minimisation of the equivalent measure of cumulative regret. The regret is the difference between the cumulative rewards of the SSP $\mathcal{S}_1$ and the one that could be acquired by a policy assumed to be optimal. In our case, the optimal policy (or the oracle strategy) consists in playing for each auction $i$ the price   $q_i=x_i\,\mathbf{1}_{\left\{x_i\leq p_i\right\}}.$

We consider two baseline strategies, corresponding to two distinct state-of-the-art policies:
\begin{itemize}
	\item the Upper Confidence Bound (UCB) policy \cite{Auer2002,sebastien2012}. Under the assumption that the rewards of each arm are independent, identically distributed, and bounded, the UCB policy achieves an order-optimal upper bound on the cumulative regret;
	\item the Exponential-weight algorithm for Exploration and Exploitation (Exp3) policy \cite{Bianchi2002,sebastien2012}. Without any assumption on the possibly non-stationary sequence of rewards (except for boundedness), the Exp3 policy achieves a worst-case order-optimal upper bound on the cumulative regret.
\end{itemize}

The number of arms $J$ has a high impact on the performance of these baseline strategies. A high number of arms makes the discretization of the coefficient applied to the bid $p_i$ very precise, but slows the convergence as the average reward for each arm is learnt independently. We have used $J=100$ in the experiments.

\subsection{Evaluation of the Thompson sampling strategy}\label{results}

This section compares the performance of the Thompson sampling strategy (TS) defined in Sections~\ref{sec:TS} and~\ref{models} with the performance of the two baseline strategies (UCB and Exp3) introduced in Section~\ref{models2} on the datasets $P_1$ and $P_2$.

The performance of a strategy after $n$ auctions is measured with the average reward:
\[
\frac{1}{n} \sum_{i=1}^n \mathbf{1}_{q_i \geq x_i} (p_i - q_i) \,.
\]

\begin{figure}
 \begin{minipage}[b]{.45\linewidth}
  \centering\includegraphics[width=1.0\linewidth,height=4cm]{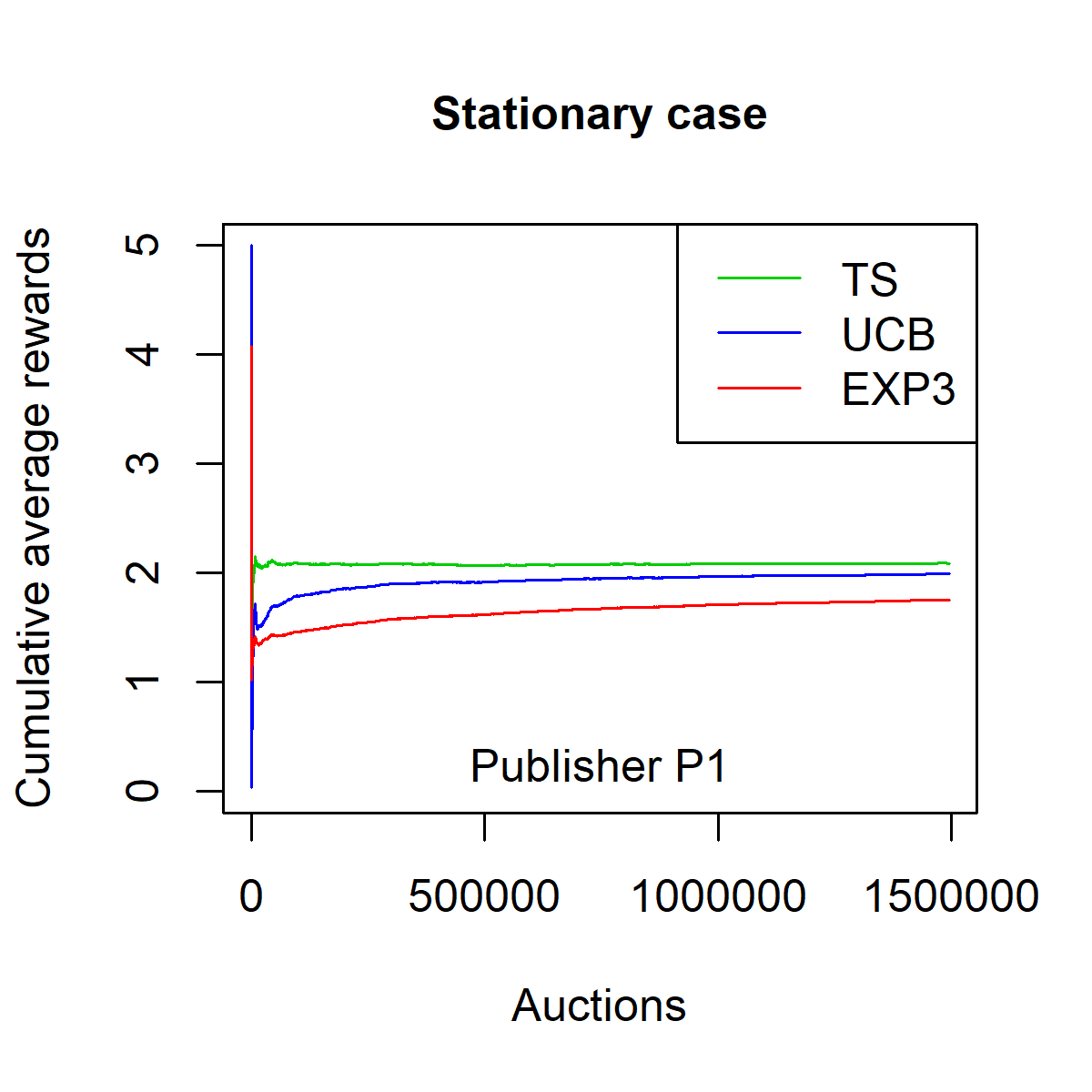}
  \caption{\footnotesize Evolution of the average rewards of TS, UCB, and Exp3 for dataset P1 (stationary environment). \label{fig:fpa_3621_shuffled_models-013}}
 \end{minipage} \hfill
 \begin{minipage}[b]{.45\linewidth}
  \centering\includegraphics[width=1.0\linewidth,height=4cm]{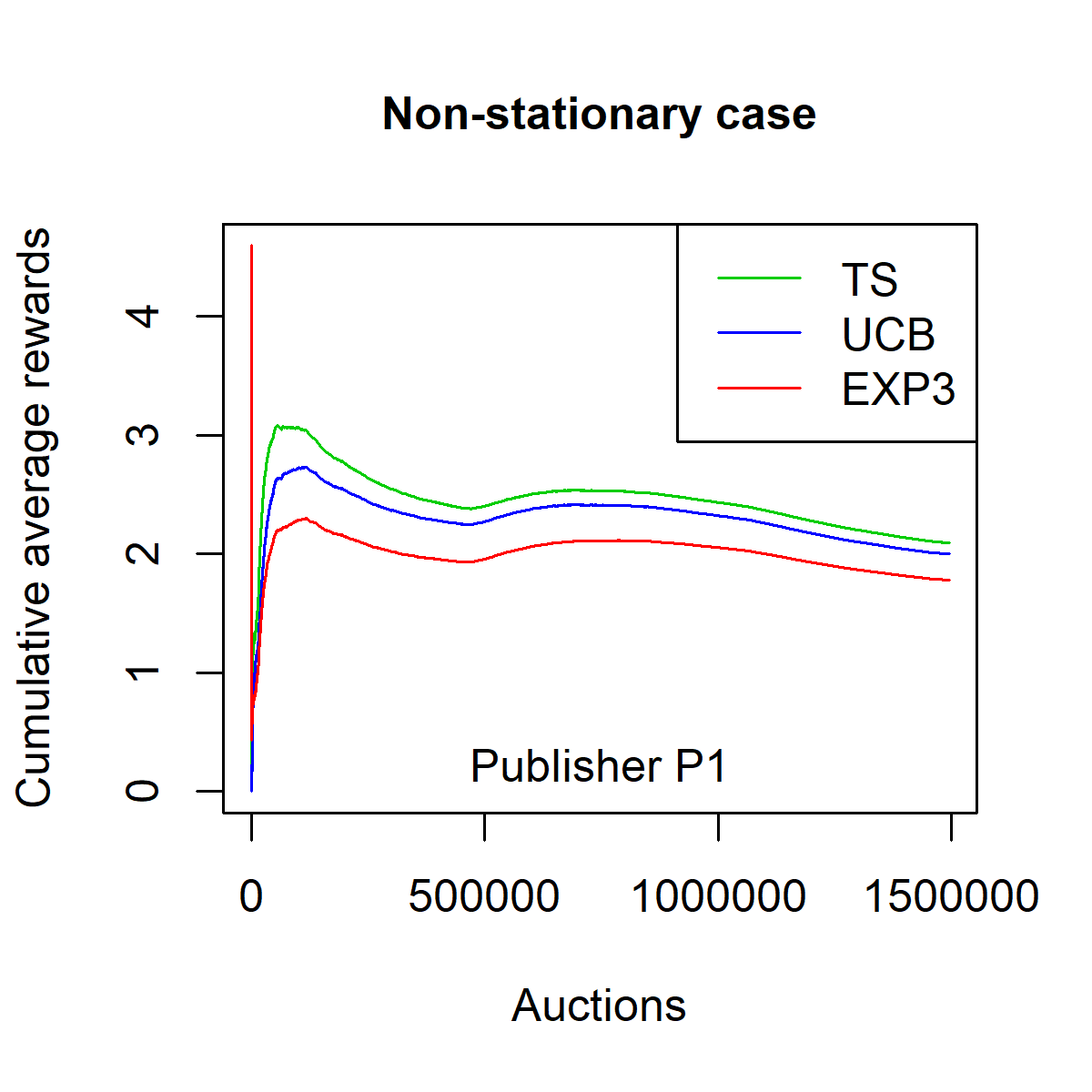}
  \caption{\footnotesize Evolution of the average rewards of TS, UCB, and Exp3 for dataset P1 (non-stationary environment). \label{fig:fpa_3621_sorted_model-013}}
 \end{minipage}
 \begin{minipage}[b]{.45\linewidth}
  \centering\includegraphics[width=1.0\linewidth,height=4cm]{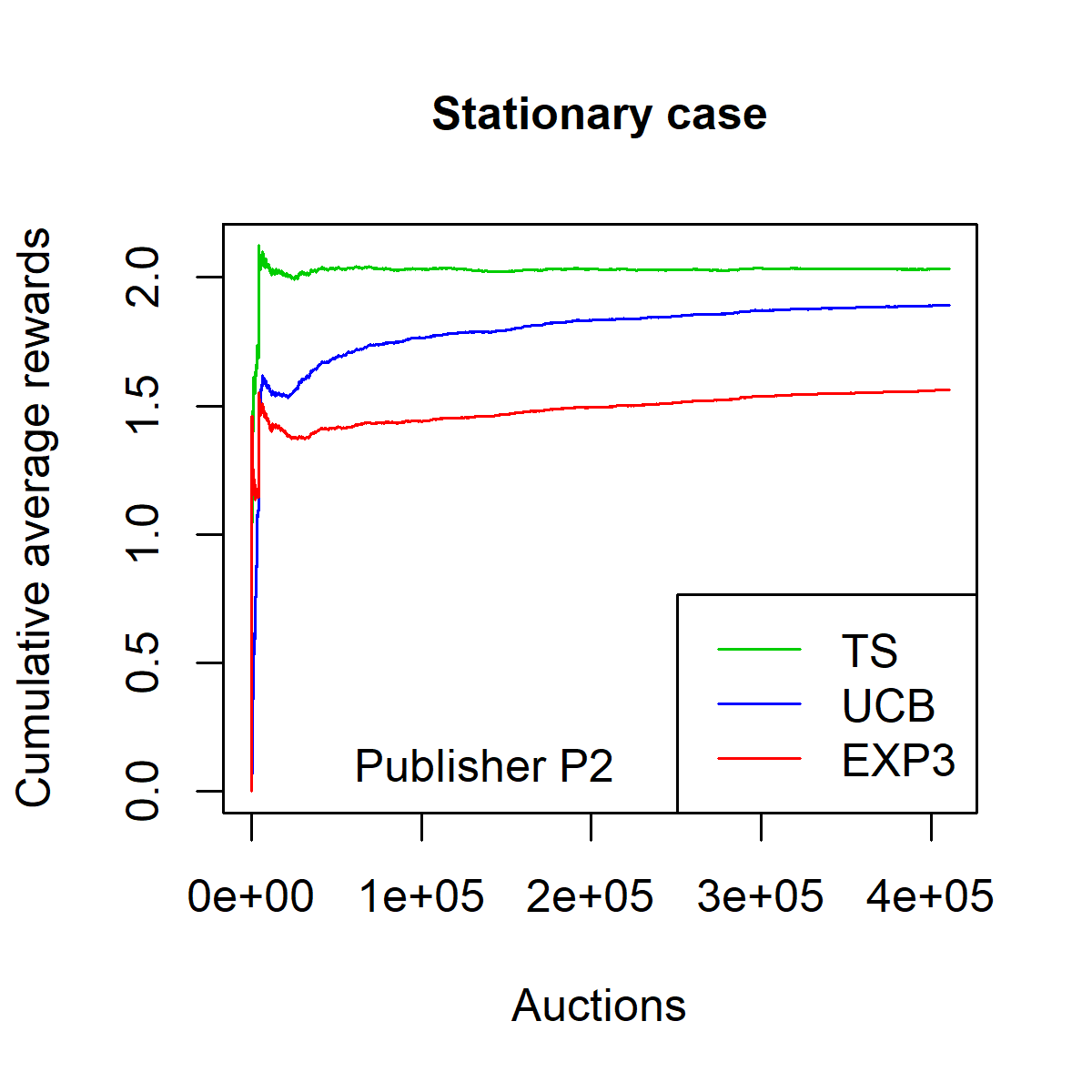}
  \caption{\footnotesize Evolution of the average rewards of TS, UCB, and Exp3 for dataset P2 (stationary environment). \label{fig:fpa_1324_shuffled_models-013}}
 \end{minipage} \hfill
 \begin{minipage}[b]{.45\linewidth}
  \centering\includegraphics[width=1.0\linewidth,height=4cm]{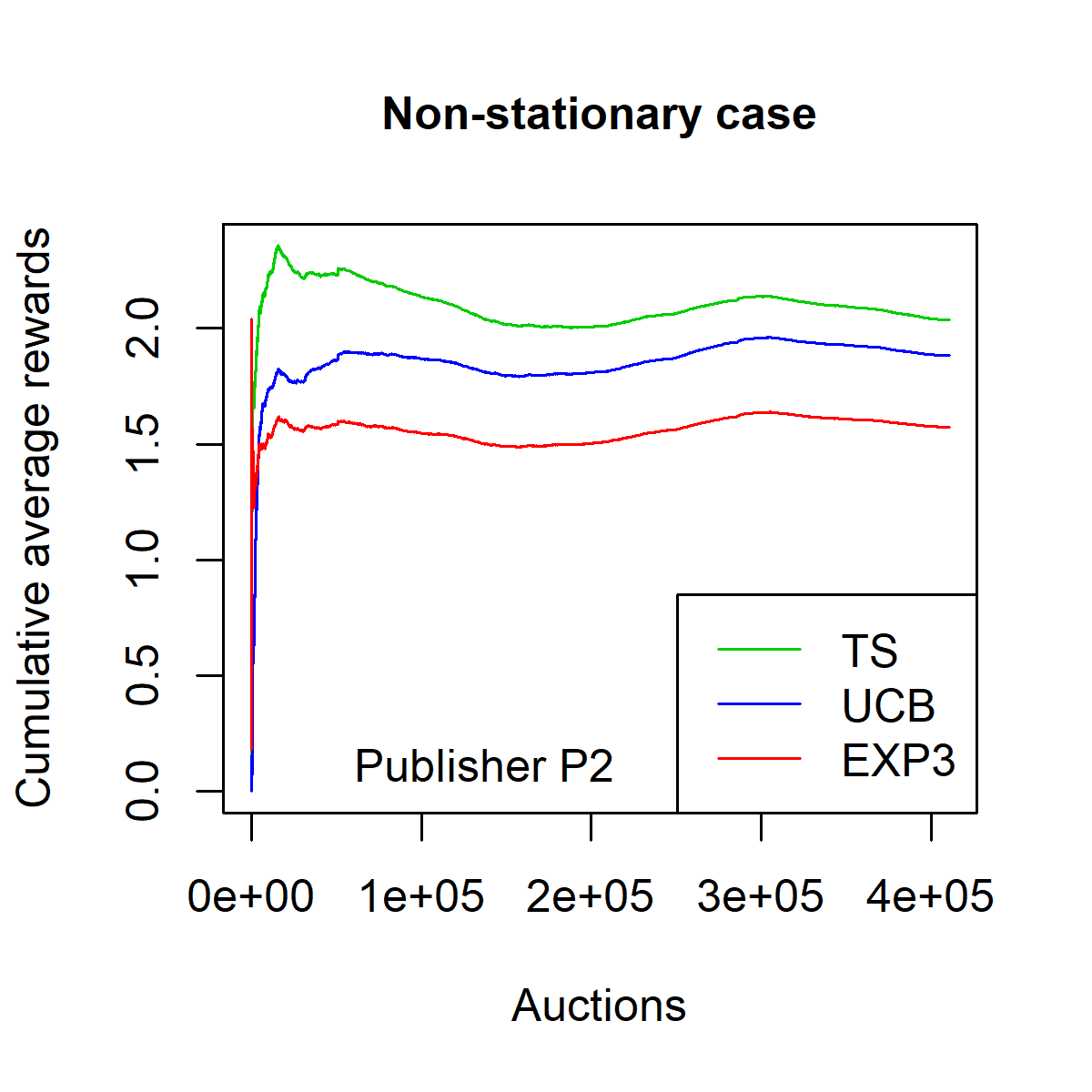}
  \caption{\footnotesize Evolution of the average rewards of TS, UCB, and Exp3 for dataset P2 (non-stationary environment) \label{fig:fpa_1324_sorted_models-013}}
 \end{minipage}
\end{figure}

Figures~\ref{fig:fpa_3621_shuffled_models-013}-\ref{fig:fpa_3621_sorted_model-013} plot the average reward as a function of $n$ on dataset $P_1$, in a stationary environment (i.e. on the shuffled dataset) and in a non-stationary environment (i.e. on the ordered dataset). Figures~\ref{fig:fpa_1324_shuffled_models-013}-\ref{fig:fpa_1324_sorted_models-013} plot the same results on dataset $P_2$.

The TS strategy clearly outperforms the baseline strategies EXP3 and UCB in both stationary and non-stationary environments. Moreover, one can observe that the convergence of the TS strategy is faster than that of the EXP3 and the UCB strategies. This convergence speed is expressed in terms of the smallest number of auctions needed by the strategy to reach the overall average reward on the whole dataset.

On the dataset $P_1$, the average reward with TS strategy is $2.0888$ for the stationary case and $2.0937$ for the non-stationary case. The corresponding success rates (i.e. the share of auctions won $n^{-1} \sum_{i=1}^n \mathbf{1}_{q_i \geq x_i}$) are $32.19\%$ and $32.09\%$ respectively.

On the dataset $P_2$, the average reward with TS strategy is $2.0312$ for the stationary case and $2.0342$ for the non-stationary case. The corresponding success rates are $29.36\%$ and $29.10\%$ respectively.

\subsection{Advantages of the Thompson sampling strategy}

The main advantage of the Thompson sampling strategy introduced in this paper is that it relies on a random modeling of the highest bid among other SSPs $x_i$, which is the unknown variable. Then, the revenue function of the problem is introduced explicitly to determine an optimal bid in each auction. In the strategies EXP3 and UCB, the rewards corresponding to each arm are learnt independently whereas they are highly correlated because they derive from a common revenue function.

In addition, as argued above, the use of a particle filter within Thompson sampling permits to handle elegantly a parameter drift, a problem which is still under investigation for classical bandit algorithms. We ran experiments using the non-stationary bandits algorithms of~\cite{Garivier2011}, but the results were not better than plain UCB strategies. In contrast, the algorithm proposed above significantly outperforms the classical approaches. 

The price for this improvement is an increased computational cost (proportional to the number of particles), and the presence of an additional  parameter $\epsilon$ which controls the intensity of the drift. It must be chosen so as to reach a good tradeoff between accuracy of the discrete approximation and the adaptation to the parameter drift. Experiments show, however, that even a very rough choice of $\epsilon$ does lead to good performance, and that over-estimating the drift intensity has little impact.

\subsection{Discussion on the parameters of the Thompson sampling strategy}

\subsubsection{Choice of the contexts}

As precised in Section~\ref{contexts}, the number of contexts has a high impact on the performance of the strategy and should be chosen carefully.


In the experiments presented in this paper, we have defined the context as the closing price of the internal auction run by the SSP $\mathcal{S}_1$. This definition of the context is intuitively a good choice, as the result of the internal auction measures the value of the ad space being sold according to the advertisers bidding in this auction. This value is probably highly correlated with the bids of other SSPs for this ad space.

The definition of the contexts could be improved by using characteristics of the ad placement or of the internet user. Experiments show that defining the context as the ad placement does not improve the results.

\subsubsection{Choice of the parametric distribution}

We chose the lognormal distribution to model the highest bid $x_i$ among the other SSPs both because it is frequently used in practice for online auctions and it fitted our datasets reasonably well. However, when the number of other SSPs is sufficiently large, using the generalized extreme value distributions or the generalized Pareto distributions \cite{Embrechts97,Coles01} might be more relevant.

Some preliminary studies we conducted show that Fr\'{e}chet distributions fit well the sample maxima of the bids $x_i$ within each context. The reason is that such probability distributions are stable and relevant to model and to track the extreme  values (sample maxima or peaks over threshold) of independent and identically distributed  random variables, whatever the behavior of their tail distributions. In such situations, the associated Thompson sampling strategy could yield even higher cumulative revenues.



\subsubsection{Computation time}

We have measured the running time (the CPU response time) of the TS strategy using standard computer ($\mu$P $2.8$GHz, RAM $8$GB). Updating the full distribution model and estimating the optimal price $q_i$ for an auction $i$ requires about $0.14$ms. This running time is below the limit of $1$ms at which the optimal price must be decided.

Note that the running time is strongly related to the parametric probability distribution modeling the highest bid among other SSPs $x_i$ and to the number of particles $K$ used to approximate the corresponding posterior distributions. 

\section{Conclusion and future work}

We have formalized the problem of optimizing the sequence of bids of a given SSP as a contextual stochastic bandit problem. This problem is tackled using the Thompson sampling algorithm, which relies on a bayesian parametric estimation of the distribution of the highest bid among other SSPs.
The distribution of the highest bid among other SSPs is approximated with a particle filtering approach. It provides a very efficient way to sequentially update the distribution and sample from it to apply the Thompson sampling algorithm.

The results obtained on datasets artificially built from real RTB auctions show that the Thompson sampling strategy outperforms other bandit approaches for this problem. Also, the estimation of the optimal bid for each impression is fast enough and the strategy can be used in real conditions where a bid prediction must be performed in a few milliseconds. This strategy is currently being developed to be deployed on thousands of web publishers worldwide.

The particle filtering models naturally the non-stationarity of the bid distributions through the hypothesis $p(\theta_{c,k,t}|\theta_{c,k,t-1})$. This hypothesis should be linked to the non-stationarity of the distributions, as decreasing its standard deviation (named $\epsilon$ in the paper) enables to forget past observations faster.

In the approach described here, the contexts are modeled independently. The learning speed of the algorithm could be increased by taking into account the correlations between the contexts. In particular, these correlations may be very high when the context is defined by a continuous variable. This point may lead to improvements in the strategy.

Finally, we are planning to explore further how the performance of the strategy depends on the parametric distribution used to model the highest bid among other SSPs $x_i$.

\bigskip

\textbf{Aknowledgment}: This work was partially funded by the French Government under the grant \verb?<ANR-13-CORD-0020>? (ALICIA Project).

\bibliographystyle{ACM-Reference-Format}
\balance
\bibliography{header_bidding}


\begin{thebibliography}{34}


\ifx \showCODEN    \undefined \def \showCODEN     #1{\unskip}     \fi
\ifx \showDOI      \undefined \def \showDOI       #1{#1}\fi
\ifx \showISBNx    \undefined \def \showISBNx     #1{\unskip}     \fi
\ifx \showISBNxiii \undefined \def \showISBNxiii  #1{\unskip}     \fi
\ifx \showISSN     \undefined \def \showISSN      #1{\unskip}     \fi
\ifx \showLCCN     \undefined \def \showLCCN      #1{\unskip}     \fi
\ifx \shownote     \undefined \def \shownote      #1{#1}          \fi
\ifx \showarticletitle \undefined \def \showarticletitle #1{#1}   \fi
\ifx \showURL      \undefined \def \showURL       {\relax}        \fi
\providecommand\bibfield[2]{#2}
\providecommand\bibinfo[2]{#2}
\providecommand\natexlab[1]{#1}
\providecommand\showeprint[2][]{arXiv:#2}

\bibitem[\protect\citeauthoryear{Auer, Cesa-Bianchi, and Fischer}{Auer
  et~al\mbox{.}}{2002a}]%
        {Auer2002}
\bibfield{author}{\bibinfo{person}{Peter Auer}, \bibinfo{person}{Nicol\`{o}
  Cesa-Bianchi}, {and} \bibinfo{person}{Paul Fischer}.}
  \bibinfo{year}{2002}\natexlab{a}.
\newblock \showarticletitle{Finite-time Analysis of the Multiarmed Bandit
  Problem}.
\newblock \bibinfo{journal}{\emph{Mach. Learn.}} \bibinfo{volume}{47},
  \bibinfo{number}{2-3} (\bibinfo{date}{May} \bibinfo{year}{2002}),
  \bibinfo{pages}{235--256}.
\newblock
\showISSN{0885-6125}
\urldef\tempurl%
\url{https://doi.org/10.1023/A:1013689704352}
\showDOI{\tempurl}


\bibitem[\protect\citeauthoryear{Auer, Cesa-Bianchi, Freund, and Schapire}{Auer
  et~al\mbox{.}}{2002b}]%
        {Bianchi2002}
\bibfield{author}{\bibinfo{person}{Peter Auer}, \bibinfo{person}{Nicol\`{o}
  Cesa-Bianchi}, \bibinfo{person}{Yoav Freund}, {and}
  \bibinfo{person}{Robert~E. Schapire}.} \bibinfo{year}{2002}\natexlab{b}.
\newblock \showarticletitle{The Nonstochastic Multiarmed Bandit Problem}.
\newblock \bibinfo{journal}{\emph{SIAM J. Comput.}} \bibinfo{volume}{32},
  \bibinfo{number}{1} (\bibinfo{year}{2002}), \bibinfo{pages}{48--77}.
\newblock
\urldef\tempurl%
\url{https://doi.org/10.1137/S0097539701398375}
\showDOI{\tempurl}
\showeprint{https://doi.org/10.1137/S0097539701398375}


\bibitem[\protect\citeauthoryear{Bubeck and Cesa{-}Bianchi}{Bubeck and
  Cesa{-}Bianchi}{2012}]%
        {sebastien2012}
\bibfield{author}{\bibinfo{person}{S{\'{e}}bastien Bubeck} {and}
  \bibinfo{person}{Nicol{\`{o}} Cesa{-}Bianchi}.}
  \bibinfo{year}{2012}\natexlab{}.
\newblock \showarticletitle{Regret Analysis of Stochastic and Nonstochastic
  Multi-armed Bandit Problems}.
\newblock \bibinfo{journal}{\emph{CoRR}}  \bibinfo{volume}{abs/1204.5721}
  (\bibinfo{year}{2012}).
\newblock
\urldef\tempurl%
\url{http://arxiv.org/abs/1204.5721}
\showURL{%
\tempurl}


\bibitem[\protect\citeauthoryear{Capp{\'e}, Moulines, and Ryden}{Capp{\'e}
  et~al\mbox{.}}{2005}]%
        {Cappe05}
\bibfield{author}{\bibinfo{person}{Olivier Capp{\'e}}, \bibinfo{person}{Eric
  Moulines}, {and} \bibinfo{person}{Tobias Ryden}.}
  \bibinfo{year}{2005}\natexlab{}.
\newblock \bibinfo{booktitle}{\emph{Inference in Hidden Markov Models (Springer
  Series in Statistics)}}.
\newblock \bibinfo{publisher}{Springer-Verlag New York, Inc.},
  \bibinfo{address}{Secaucus, NJ, USA}.
\newblock
\showISBNx{0387402640}


\bibitem[\protect\citeauthoryear{Cesa-Bianchi, Gentile, and
  Mansour}{Cesa-Bianchi et~al\mbox{.}}{2015}]%
        {CeGeMa-15-ReservePriceOptimization}
\bibfield{author}{\bibinfo{person}{N. Cesa-Bianchi}, \bibinfo{person}{C.
  Gentile}, {and} \bibinfo{person}{Y. Mansour}.}
  \bibinfo{year}{2015}\natexlab{}.
\newblock \showarticletitle{Regret minimization for reserve prices in
  second-price auctions}.
\newblock \bibinfo{journal}{\emph{IEEE Trans. Inform. Theory}}
  \bibinfo{volume}{61}, \bibinfo{number}{1} (\bibinfo{year}{2015}),
  \bibinfo{pages}{549--564}.
\newblock


\bibitem[\protect\citeauthoryear{Cesa-Bianchi and Lugosi}{Cesa-Bianchi and
  Lugosi}{2006}]%
        {PLG06}
\bibfield{author}{\bibinfo{person}{Nicolo Cesa-Bianchi} {and}
  \bibinfo{person}{Gabor Lugosi}.} \bibinfo{year}{2006}\natexlab{}.
\newblock \bibinfo{booktitle}{\emph{Prediction, Learning, and Games}}.
\newblock \bibinfo{publisher}{Cambridge University Press},
  \bibinfo{address}{New York, NY, USA}.
\newblock
\showISBNx{0521841089}


\bibitem[\protect\citeauthoryear{Coles}{Coles}{2001}]%
        {Coles01}
\bibfield{author}{\bibinfo{person}{S.~G. Coles}.}
  \bibinfo{year}{2001}\natexlab{}.
\newblock \bibinfo{booktitle}{\emph{An introduction to statistical modeling of
  extreme values}}.
\newblock \bibinfo{publisher}{Springer Series in Statistics}.
\newblock


\bibitem[\protect\citeauthoryear{Crisan and Doucet}{Crisan and Doucet}{2000}]%
        {Crisan2000}
\bibfield{author}{\bibinfo{person}{Dan Crisan} {and} \bibinfo{person}{Arnaud
  Doucet}.} \bibinfo{year}{2000}\natexlab{}.
\newblock \bibinfo{booktitle}{\emph{Convergence of Sequential Monte Carlo
  Methods}}.
\newblock \bibinfo{type}{{T}echnical {R}eport}.
\newblock


\bibitem[\protect\citeauthoryear{Crisan and Doucet}{Crisan and Doucet}{2002}]%
        {Crisan2002}
\bibfield{author}{\bibinfo{person}{Dan Crisan} {and} \bibinfo{person}{Arnaud
  Doucet}.} \bibinfo{year}{2002}\natexlab{}.
\newblock \bibinfo{title}{A Survey of Convergence Results on Particle Filtering
  Methods for Practitioners}.
\newblock   (\bibinfo{year}{2002}).
\newblock


\bibitem[\protect\citeauthoryear{Deng, Goldberg, Tang, and Zhang}{Deng
  et~al\mbox{.}}{2014}]%
        {intermediary_bayesian}
\bibfield{author}{\bibinfo{person}{Xiaotie Deng}, \bibinfo{person}{Paul~W.
  Goldberg}, \bibinfo{person}{Bo Tang}, {and} \bibinfo{person}{Jinshan Zhang}.}
  \bibinfo{year}{2014}\natexlab{}.
\newblock \showarticletitle{Revenue maximization in a Bayesian double auction
  market}.
\newblock \bibinfo{journal}{\emph{Theor. Comput. Sci.}}  \bibinfo{volume}{539}
  (\bibinfo{year}{2014}), \bibinfo{pages}{1--12}.
\newblock
\urldef\tempurl%
\url{https://doi.org/10.1016/j.tcs.2014.04.013}
\showDOI{\tempurl}


\bibitem[\protect\citeauthoryear{Douc, Garivier, Moulines, and Olsson}{Douc
  et~al\mbox{.}}{2011}]%
        {Douc2012}
\bibfield{author}{\bibinfo{person}{Randal Douc}, \bibinfo{person}{Aurélien
  Garivier}, \bibinfo{person}{Eric Moulines}, {and} \bibinfo{person}{Jimmy
  Olsson}.} \bibinfo{year}{2011}\natexlab{}.
\newblock \showarticletitle{Sequential Monte Carlo smoothing for general state
  space hidden Markov models}.
\newblock \bibinfo{journal}{\emph{Ann. Appl. Probab.}} \bibinfo{volume}{21},
  \bibinfo{number}{6} (\bibinfo{date}{12} \bibinfo{year}{2011}),
  \bibinfo{pages}{2109--2145}.
\newblock
\urldef\tempurl%
\url{https://doi.org/10.1214/10-AAP735}
\showDOI{\tempurl}


\bibitem[\protect\citeauthoryear{Doucet, de~Freitas, and Gordon}{Doucet
  et~al\mbox{.}}{2001}]%
        {Doucet2001}
\bibfield{editor}{\bibinfo{person}{Arnaud Doucet}, \bibinfo{person}{Nando de
  Freitas}, {and} \bibinfo{person}{Neil Gordon}} (Eds.).
  \bibinfo{year}{2001}\natexlab{}.
\newblock \bibinfo{booktitle}{\emph{Sequential Monte Carlo Methods in
  Practice}}.
\newblock \bibinfo{publisher}{Springer}.
\newblock


\bibitem[\protect\citeauthoryear{Embrechts, Kl{\"u}ppelberg, and
  Mikosch}{Embrechts et~al\mbox{.}}{1997}]%
        {Embrechts97}
\bibfield{author}{\bibinfo{person}{P. Embrechts}, \bibinfo{person}{C.
  Kl{\"u}ppelberg}, {and} \bibinfo{person}{T. Mikosch}.}
  \bibinfo{year}{1997}\natexlab{}.
\newblock \bibinfo{booktitle}{\emph{Modelling Extremal Events for Insurance and
  Finance}}.
\newblock \bibinfo{publisher}{Springer-Verlag, Berlin}.
\newblock


\bibitem[\protect\citeauthoryear{Feldman, Mirrokni, Muthukrishnan, and
  Pai}{Feldman et~al\mbox{.}}{2010}]%
        {auctions_intermediaries}
\bibfield{author}{\bibinfo{person}{Jon Feldman}, \bibinfo{person}{Vahab
  Mirrokni}, \bibinfo{person}{S. Muthukrishnan}, {and}
  \bibinfo{person}{Mallesh~M. Pai}.} \bibinfo{year}{2010}\natexlab{}.
\newblock \showarticletitle{Auctions with Intermediaries: Extended Abstract}.
  In \bibinfo{booktitle}{\emph{Proceedings of the 11th ACM Conference on
  Electronic Commerce}} \emph{(\bibinfo{series}{EC '10})}.
  \bibinfo{publisher}{ACM}, \bibinfo{address}{New York, NY, USA},
  \bibinfo{pages}{23--32}.
\newblock
\showISBNx{978-1-60558-822-3}
\urldef\tempurl%
\url{https://doi.org/10.1145/1807342.1807346}
\showDOI{\tempurl}


\bibitem[\protect\citeauthoryear{{Fernandez-Tapia}, {Gu{\'e}ant}, and
  {Lasry}}{{Fernandez-Tapia} et~al\mbox{.}}{2015}]%
        {optimal_bidding}
\bibfield{author}{\bibinfo{person}{J. {Fernandez-Tapia}}, \bibinfo{person}{O.
  {Gu{\'e}ant}}, {and} \bibinfo{person}{J.-M. {Lasry}}.}
  \bibinfo{year}{2015}\natexlab{}.
\newblock \showarticletitle{{Optimal Real-Time Bidding Strategies}}.
\newblock \bibinfo{journal}{\emph{ArXiv e-prints}} (\bibinfo{date}{Nov.}
  \bibinfo{year}{2015}).
\newblock
\showeprint[arxiv]{math.OC/1511.08409}


\bibitem[\protect\citeauthoryear{Garivier and Moulines}{Garivier and
  Moulines}{2011}]%
        {Garivier2011}
\bibfield{author}{\bibinfo{person}{Aur{\'e}lien Garivier} {and}
  \bibinfo{person}{Eric Moulines}.} \bibinfo{year}{2011}\natexlab{}.
\newblock \bibinfo{booktitle}{\emph{On Upper-Confidence Bound Policies for
  Switching Bandit Problems}}.
\newblock \bibinfo{publisher}{Springer Berlin Heidelberg},
  \bibinfo{address}{Berlin, Heidelberg}, \bibinfo{pages}{174--188}.
\newblock
\showISBNx{978-3-642-24412-4}
\urldef\tempurl%
\url{https://doi.org/10.1007/978-3-642-24412-4_16}
\showDOI{\tempurl}


\bibitem[\protect\citeauthoryear{Gomes and Mirrokni}{Gomes and
  Mirrokni}{2014}]%
        {adx}
\bibfield{author}{\bibinfo{person}{Renato Gomes} {and}
  \bibinfo{person}{Vahab~S. Mirrokni}.} \bibinfo{year}{2014}\natexlab{}.
\newblock \showarticletitle{Optimal revenue-sharing double auctions with
  applications to ad exchanges}. In \bibinfo{booktitle}{\emph{23rd
  International World Wide Web Conference, {WWW} '14, Seoul, Republic of Korea,
  April 7-11, 2014}}. \bibinfo{pages}{19--28}.
\newblock
\urldef\tempurl%
\url{https://doi.org/10.1145/2566486.2568029}
\showDOI{\tempurl}


\bibitem[\protect\citeauthoryear{Heidari, Mahdian, Syed, Vassilvitskii, and
  Yazdanbod}{Heidari et~al\mbox{.}}{2016}]%
        {HeidariETAL-16-PricingLowRegretSeller}
\bibfield{author}{\bibinfo{person}{H. Heidari}, \bibinfo{person}{M. Mahdian},
  \bibinfo{person}{U. Syed}, \bibinfo{person}{S. Vassilvitskii}, {and}
  \bibinfo{person}{S. Yazdanbod}.} \bibinfo{year}{2016}\natexlab{}.
\newblock \showarticletitle{Pricing a low-regret seller}, In
  \bibinfo{booktitle}{Proceedings of The 33rd International Conference on
  Machine Learning (ICML 2016)}.
\newblock \bibinfo{journal}{\emph{Proceedings of The 33rd International
  Conference on Machine Learning (ICML'16)}}, \bibinfo{pages}{2559--2567}.
\newblock


\bibitem[\protect\citeauthoryear{{Kaufmann}, {Korda}, and {Munos}}{{Kaufmann}
  et~al\mbox{.}}{2012}]%
        {Kaufmann2012}
\bibfield{author}{\bibinfo{person}{E. {Kaufmann}}, \bibinfo{person}{N.
  {Korda}}, {and} \bibinfo{person}{R. {Munos}}.}
  \bibinfo{year}{2012}\natexlab{}.
\newblock \showarticletitle{{Thompson Sampling: An Asymptotically Optimal
  Finite Time Analysis}}.
\newblock \bibinfo{journal}{\emph{ArXiv e-prints}} (\bibinfo{date}{May}
  \bibinfo{year}{2012}).
\newblock
\showeprint[arxiv]{stat.ML/1205.4217}


\bibitem[\protect\citeauthoryear{Kleinberg and Leighton}{Kleinberg and
  Leighton}{2003}]%
        {KlLe-03-PostedPriceAuctions}
\bibfield{author}{\bibinfo{person}{R. Kleinberg} {and} \bibinfo{person}{T.
  Leighton}.} \bibinfo{year}{2003}\natexlab{}.
\newblock \showarticletitle{The value of knowing a demand curve: bounds on
  regret for on-line posted-price auctions}. In
  \bibinfo{booktitle}{\emph{Proceedings of the 44th IEEE Symposium on
  Foundations of Computer Science (FOCS 2003)}}. \bibinfo{pages}{594--605}.
\newblock
\newblock
\shownote{Full version avalaible at
  https://www.cs.cornell.edu/~rdk/papers/oppa.pdf.}


\bibitem[\protect\citeauthoryear{{Korda}, {Kaufmann}, and {Munos}}{{Korda}
  et~al\mbox{.}}{2013}]%
        {Kaufmann2013}
\bibfield{author}{\bibinfo{person}{N. {Korda}}, \bibinfo{person}{E.
  {Kaufmann}}, {and} \bibinfo{person}{R. {Munos}}.}
  \bibinfo{year}{2013}\natexlab{}.
\newblock \showarticletitle{{Thompson Sampling for 1-Dimensional Exponential
  Family Bandits}}.
\newblock \bibinfo{journal}{\emph{ArXiv e-prints}} (\bibinfo{date}{July}
  \bibinfo{year}{2013}).
\newblock
\showeprint[arxiv]{stat.ML/1307.3400}


\bibitem[\protect\citeauthoryear{{Lee}, {Jalali}, and {Dasdan}}{{Lee}
  et~al\mbox{.}}{2013}]%
        {optimal_bidding_3}
\bibfield{author}{\bibinfo{person}{K.-C. {Lee}}, \bibinfo{person}{A. {Jalali}},
  {and} \bibinfo{person}{A. {Dasdan}}.} \bibinfo{year}{2013}\natexlab{}.
\newblock \showarticletitle{{Real Time Bid Optimization with Smooth Budget
  Delivery in Online Advertising}}.
\newblock \bibinfo{journal}{\emph{ArXiv e-prints}} (\bibinfo{date}{May}
  \bibinfo{year}{2013}).
\newblock
\showeprint[arxiv]{cs.GT/1305.3011}


\bibitem[\protect\citeauthoryear{{Leike}, {Lattimore}, {Orseau}, and
  {Hutter}}{{Leike} et~al\mbox{.}}{2016}]%
        {Leike2016}
\bibfield{author}{\bibinfo{person}{J. {Leike}}, \bibinfo{person}{T.
  {Lattimore}}, \bibinfo{person}{L. {Orseau}}, {and} \bibinfo{person}{M.
  {Hutter}}.} \bibinfo{year}{2016}\natexlab{}.
\newblock \showarticletitle{{Thompson Sampling is Asymptotically Optimal in
  General Environments}}.
\newblock \bibinfo{journal}{\emph{ArXiv e-prints}} (\bibinfo{date}{Feb.}
  \bibinfo{year}{2016}).
\newblock
\showeprint[arxiv]{cs.LG/1602.07905}


\bibitem[\protect\citeauthoryear{Mohri and Medina}{Mohri and Medina}{2014}]%
        {MoMe-14nips-PostedPriceAuctions}
\bibfield{author}{\bibinfo{person}{M. Mohri} {and} \bibinfo{person}{A.~M.
  Medina}.} \bibinfo{year}{2014}\natexlab{}.
\newblock \showarticletitle{Optimal regret minimization in posted-price
  auctions with strategic buyers}. In \bibinfo{booktitle}{\emph{Advances in
  Neural Information Processing Systems (NIPS'14)}}.
\newblock


\bibitem[\protect\citeauthoryear{Mohri and Medina}{Mohri and Medina}{2015}]%
        {MoMe-15nips-PostedPriceAuctions-randomvaluation}
\bibfield{author}{\bibinfo{person}{M. Mohri} {and} \bibinfo{person}{A.~M.
  Medina}.} \bibinfo{year}{2015}\natexlab{}.
\newblock \showarticletitle{Revenue optimization against strategic buyers}. In
  \bibinfo{booktitle}{\emph{Advances in Neural Information Processing Systems
  (NIPS'15)}}.
\newblock


\bibitem[\protect\citeauthoryear{Murphy}{Murphy}{2012}]%
        {murphy}
\bibfield{author}{\bibinfo{person}{Kevin~P. Murphy}.}
  \bibinfo{year}{2012}\natexlab{}.
\newblock \bibinfo{booktitle}{\emph{Machine Learning: A Probabilistic
  Perspective} (\bibinfo{edition}{6th} ed.)}.
\newblock \bibinfo{publisher}{Cambridge, MA: MIT Press}.
\newblock


\bibitem[\protect\citeauthoryear{Myerson}{Myerson}{1981}]%
        {myerson}
\bibfield{author}{\bibinfo{person}{Roger~B. Myerson}.}
  \bibinfo{year}{1981}\natexlab{}.
\newblock \showarticletitle{Optimal Auction Design}.
\newblock \bibinfo{journal}{\emph{Math. Oper. Res.}} \bibinfo{volume}{6},
  \bibinfo{number}{1} (\bibinfo{date}{Feb.} \bibinfo{year}{1981}),
  \bibinfo{pages}{58--73}.
\newblock
\showISSN{0364-765X}
\urldef\tempurl%
\url{https://doi.org/10.1287/moor.6.1.58}
\showDOI{\tempurl}


\bibitem[\protect\citeauthoryear{Myerson and Satterthwaite}{Myerson and
  Satterthwaite}{1983}]%
        {myerson_bilateral}
\bibfield{author}{\bibinfo{person}{Roger~B Myerson} {and}
  \bibinfo{person}{Mark~A Satterthwaite}.} \bibinfo{year}{1983}\natexlab{}.
\newblock \showarticletitle{Efficient mechanisms for bilateral trading}.
\newblock \bibinfo{journal}{\emph{Journal of Economic Theory}}
  \bibinfo{volume}{29}, \bibinfo{number}{2} (\bibinfo{year}{1983}),
  \bibinfo{pages}{265 -- 281}.
\newblock
\showISSN{0022-0531}
\urldef\tempurl%
\url{https://doi.org/10.1016/0022-0531(83)90048-0}
\showDOI{\tempurl}


\bibitem[\protect\citeauthoryear{Qin, Yuan, and Wang}{Qin
  et~al\mbox{.}}{2017}]%
        {adx_header_bidding}
\bibfield{author}{\bibinfo{person}{Rui Qin}, \bibinfo{person}{Yong Yuan}, {and}
  \bibinfo{person}{Fei-Yue Wang}.} \bibinfo{year}{2017}\natexlab{}.
\newblock \showarticletitle{Optimizing the Revenue for Ad Exchanges in Header
  Bidding Advertising Markets}. In \bibinfo{booktitle}{\emph{IEEE International
  Conference on Systems, Man, and Cybernetics}}.
\newblock


\bibitem[\protect\citeauthoryear{Thompson}{Thompson}{1933}]%
        {thompson}
\bibfield{author}{\bibinfo{person}{W.~R. Thompson}.}
  \bibinfo{year}{1933}\natexlab{}.
\newblock \showarticletitle{{On the Likelihood that one Unknown Probability
  Exceeds Another in View of the Evidence of Two Samples}}.
\newblock \bibinfo{journal}{\emph{Biometrika}}  \bibinfo{volume}{25}
  (\bibinfo{year}{1933}), \bibinfo{pages}{285--294}.
\newblock


\bibitem[\protect\citeauthoryear{Vidakovic}{Vidakovic}{2017}]%
        {header_bidding}
\bibfield{author}{\bibinfo{person}{Ratko Vidakovic}.}
  \bibinfo{year}{2017}\natexlab{}.
\newblock \bibinfo{title}{The Beginner's Guide To Header Bidding}.
\newblock   (\bibinfo{date}{March} \bibinfo{year}{2017}).
\newblock


\bibitem[\protect\citeauthoryear{Wang, Zhang, and Yuan}{Wang
  et~al\mbox{.}}{2017}]%
        {Wang2017}
\bibfield{author}{\bibinfo{person}{Jun Wang}, \bibinfo{person}{Weinan Zhang},
  {and} \bibinfo{person}{Shuai Yuan}.} \bibinfo{year}{2017}\natexlab{}.
\newblock \showarticletitle{Display Advertising with Real-Time Bidding (RTB)
  and Behavioural Targeting}.
\newblock \bibinfo{journal}{\emph{Foundations and Trends in Information
  Retrieval}} \bibinfo{volume}{11}, \bibinfo{number}{4-5}
  (\bibinfo{year}{2017}), \bibinfo{pages}{297--435}.
\newblock
\showISSN{1554-0669}
\urldef\tempurl%
\url{https://doi.org/10.1561/1500000049}
\showDOI{\tempurl}


\bibitem[\protect\citeauthoryear{Weed, Perchet, and Rigollet}{Weed
  et~al\mbox{.}}{2016}]%
        {WePeRi-16colt-RepeatedAuctions}
\bibfield{author}{\bibinfo{person}{J. Weed}, \bibinfo{person}{V. Perchet},
  {and} \bibinfo{person}{P. Rigollet}.} \bibinfo{year}{2016}\natexlab{}.
\newblock \showarticletitle{Online learning in repeated auctions}. In
  \bibinfo{booktitle}{\emph{Proceedings of the 29th Annual Conference on
  Learning Theory (COLT'16)}}.
\newblock


\bibitem[\protect\citeauthoryear{Zhang, Yuan, and Wang}{Zhang
  et~al\mbox{.}}{2014}]%
        {optimal_bidding_2}
\bibfield{author}{\bibinfo{person}{Weinan Zhang}, \bibinfo{person}{Shuai Yuan},
  {and} \bibinfo{person}{Jun Wang}.} \bibinfo{year}{2014}\natexlab{}.
\newblock \showarticletitle{Optimal Real-time Bidding for Display Advertising}.
  In \bibinfo{booktitle}{\emph{Proceedings of the 20th ACM SIGKDD International
  Conference on Knowledge Discovery and Data Mining}}
  \emph{(\bibinfo{series}{KDD '14})}. \bibinfo{publisher}{ACM},
  \bibinfo{address}{New York, NY, USA}, \bibinfo{pages}{1077--1086}.
\newblock
\showISBNx{978-1-4503-2956-9}
\urldef\tempurl%
\url{https://doi.org/10.1145/2623330.2623633}
\showDOI{\tempurl}


\end{thebibliography}

\end{document}